\pdfoutput=1

\documentclass[11pt]{article}

\usepackage[preprint]{acl}

\usepackage{times}
\usepackage{latexsym}
\usepackage[T1]{fontenc}
\usepackage[utf8]{inputenc}
\usepackage{microtype}
\usepackage{graphicx}
\usepackage{booktabs}
\usepackage{amsmath}
\usepackage{amssymb}
\usepackage{multirow}
\usepackage{algorithm}
\usepackage{algpseudocode}
\usepackage{xspace}
\newcommand{\chrf}{chrF++\xspace}

\newcommand{\coderepo}{https://github.com/AbdulRehmanAntall/morphaustro-tok}

\title{Does Morphological Correctness Matter for Tokenization? \\
A Scrambled Control Study in Tagalog}

\author{
  \textbf{Abdul Rehman Antall\textsuperscript{$\spadesuit$}} \quad
  \textbf{Baidya Nath Saha\textsuperscript{$\clubsuit$}} \\
  \textsuperscript{$\spadesuit$}National University of Computer and Emerging
  Sciences (FAST-NUCES), Lahore, Pakistan \\
  \textsuperscript{$\clubsuit$}Concordia University of Edmonton, Alberta, Canada \\
  \small{\texttt{l230605@lhr.nu.edu.pk} \quad
         \texttt{baidya.saha@concordia.ab.ca}}
}

\hypersetup{
  pdftitle={Does Morphological Correctness Matter for Tokenization?
            A Scrambled Control Study in Tagalog},
  pdfauthor={Abdul Rehman Antall, Baidya Nath Saha},
  pdfkeywords={tokenization, subword segmentation, morphology, Tagalog,
               machine translation, low-resource NLP}
}

\begin{document}
\maketitle

\begin{abstract}
Typology-aware tokenizers are motivated by the claim that segmentation which
respects morpheme boundaries preserves information that frequency-driven
segmentation destroys. Studies routinely establish that it helps while assuming
the boundaries must be linguistically correct. We separate them. Tagalog marks
voice with infixes (\textit{s-um-ulat} `wrote', \textit{s-in-ulat} `was
written'), so the root is not contiguous and no sequence of BPE merges can
recover it. We build MorphAustro-Tok, a tokenizer that emits the root as a
contiguous unit followed by a positional infix token, and compare it against a
corpus-matched BPE baseline and, at two of three fertility levels, against a
scrambled-morpheme control that preserves morph count, invertibility and per-type
consistency while randomising the pseudo-infix. Across 33 runs, MorphAustro-Tok
improves Tagalog machine translation over matched BPE by 0.84, 1.49 and 3.40 mean
sentence-level \chrf points. Token count does not explain the gain, since it grows
as segmentation coarsens, and neither does capacity: at the coarsest level the
scrambled arm carries more parameters and scores lower. Scrambling the analysis
moves the gain by less than between-seed dispersion at both controlled levels, so
we detect no contribution from morphological correctness at this scale.
\end{abstract}

\section{Introduction}

The case for typology-aware tokenization rests on an irreversibility argument. A
tokenizer fixes the alphabet in which a model perceives its input, and a contrast
destroyed by segmentation cannot be recovered by any amount of downstream
capacity. Where a language marks grammatical function by means that
frequency-driven merges systematically fragment, the argument predicts a cost
that no encoder can repay \citep{bostrom-durrett-2020-byte,park-etal-2021-morphology}.

Tagalog is close to the strongest available test of that argument in a
Latin-script language. Voice is marked by the infixes \textit{-um-} (actor voice)
and \textit{-in-} (patient voice), inserted after the root-initial onset, so that
\textit{sulat} `write' becomes \textit{s-um-ulat} `wrote' and \textit{s-in-ulat}
`was written' \citep{schachter-otanes-1972-tagalog,kroeger-1993-phrase}. The root
is discontinuous in the surface string. A BPE merge joins adjacent symbols
\citep{sennrich-etal-2016-neural}, so no merge sequence can represent
\texttt{sulat} as a unit inside \texttt{sumulat}, and the paradigm fragments into
unrelated whole-word types. This is the same class of problem that root-and-pattern
morphology poses for Semitic, arriving inside an otherwise ordinary Latin-script
word.

If the irreversibility argument is right, a tokenizer that recovers the root
should help. Our results say that it does, and they also say that the recovery
does not have to be correct.

That distinction is the contribution of this paper. A tokenizer intervention has
at least two separable properties: it changes \textit{where} the cuts fall, and it
changes \textit{how consistently} a given word type is cut. A design that varies
both at once, as ours does and as the interventions in this literature generally
do, cannot attribute an observed gain to either property. A control that holds
granularity and consistency fixed while randomising linguistic content can at
least remove one of them from the explanation, and running that control is what
separates a positive result from an attributable
one.\footnote{Code, per-run records and the gate log are available at
\url{\coderepo}.}

\paragraph{Contributions.} We state up front what level of evidence supports each
claim, since the claims differ substantially in strength.

\begin{enumerate}
\setlength{\itemsep}{1pt}
\item \textbf{MorphAustro-Tok}, a two-stage tokenizer in which an infix-aware
pre-segmentation emits the Tagalog root as a contiguous unit followed by a
positional infix token, and a constrained BPE learns merges that cannot cross a
morph boundary (Algorithm~\ref{alg:stage1}).
\item A \textbf{scrambled-morpheme control} that fires on exactly the word types
the real analyzer fires on and preserves morph count, invertibility and per-type
consistency while destroying the intended morphological alignment
(Algorithm~\ref{alg:scramble}).
\item Evidence that the machine-translation gain over matched BPE is
\textbf{not explained by token count or by vocabulary-induced capacity}: it grows
monotonically as segmentation coarsens, and at the coarsest level the arm with
more parameters scores lower (\S\ref{sec:confounds}).
\item \textbf{No detectable contribution from morphological correctness} at this
scale. Against the scrambled control the difference is $-0.055$ \chrf at the
central fertility level and $+0.661$ at the coarsest, opposite in sign and both
far below between-seed dispersion (\S\ref{sec:scramble-results}).
\item One asymmetric result that survives both uncertainty floors: on named
entity recognition, scrambling \textbf{costs} 0.023 F1 relative to the real
analysis, while the real analysis does not beat BPE (\S\ref{sec:ner}).
\end{enumerate}

We do not claim that morphological correctness is inert, only that at this scale,
with these tasks, we could not detect a contribution from it at the resolution our
example-level and seed-level variation allows.

\section{Related work}
\label{sec:related}

\paragraph{The tokenizer stage.} Subword segmentation determines what a model can
represent, and tokenizer quality tracks downstream quality
\citep{rust-etal-2021-good,gowda-may-2020-finding}. Work on morphologically rich
languages has reported that morphology-aware segmentation improves downstream
results and has attributed the improvement to morphological validity
\citep{bostrom-durrett-2020-byte,hofmann-etal-2022-embarrassingly,park-etal-2021-morphology},
rarely testing that attribution against a granularity-matched control. A tokenizer
that changes segmentation \textit{consistency} alongside \textit{correctness}
leaves the two confounded, and \citet{schmidt-etal-2024-tokenization} make the
related point that compression alone does not explain what tokenizers contribute.
We take that argument further by removing correctness while holding consistency
and granularity fixed.

\paragraph{Infixation and symmetrical voice in Philippine languages.} Tagalog
exhibits a symmetrical voice system in which the semantic role of the pivot
argument is determined by verbal morphology
\citep{schachter-otanes-1972-tagalog,kroeger-1993-phrase,himmelmann-2005-austronesian}.
Two of the productive markers are infixes. Because an infix is inserted inside the
root rather than concatenated to it, the root is not a contiguous substring, and
the standard subword machinery, which operates by merging adjacent units, is
structurally unable to isolate it. Unsupervised segmenters such as Morfessor
\citep{smit-etal-2014-morfessor} inherit the restriction, since they too model
words as concatenations of contiguous morphs.

\paragraph{Numeral classifiers in Malayic.} Indonesian requires a classifier
between a numeral and a counted noun (\textit{tiga ekor kucing} `three cats'),
and its most frequent realisation fuses the numeral `one' with the classifier
into one orthographic word (\textit{seekor kucing} `one cat')
\citep{sneddon-etal-2010-indonesian}.

\paragraph{Evaluation resources.} FLORES+ provides parallel evaluation across
Austronesian varieties \citep{goyal-etal-2022-flores,nllb-2024-scaling}, and
WikiANN provides named-entity annotation
\citep{pan-etal-2017-cross,rahimi-etal-2019-massively}. Training bitext for
Tagalog is scarce \citep{baliber-etal-2020-bridging,lovenia-etal-2024-seacrowd}:
OPUS-100 \citep{zhang-etal-2020-improving} contains no Tagalog configuration, and
the mined NLLB corpus is distributed in a form no longer loadable by current
tooling. We therefore fetch ParaCrawl v9 \citep{banon-etal-2020-paracrawl}
directly from OPUS \citep{tiedemann-2012-parallel}, which keeps the two arms on a
common source.

\section{Methodology}
\label{sec:method}

\subsection{Hypotheses and the decisive criterion}

We test one hypothesis and one anti-hypothesis, both fixed before the runs were
executed. H1 states that a morpheme-boundary constraint on BPE improves
downstream quality for Tagalog relative to a corpus-matched BPE baseline. The
anti-hypothesis states that any observed gain reduces to compression and
disappears once tokens per character are held equal. A decisive criterion was
fixed alongside them: if performance survives a scramble of the morphological
analysis, correctness of the analysis is not what produced the gain. The
Indonesian classifier arm is an exploratory feasibility check rather than a
hypothesis test (\S\ref{sec:gating}). The study was not preregistered, and we use ``fixed before
the runs'' in that narrower sense.

\subsection{MorphAustro-Tok}
\label{sec:morphaustro}

\begin{algorithm}[t]
\small
\caption{Stage-1 segmentation of one word}
\label{alg:stage1}
\begin{algorithmic}[1]
\Require word $w$; root-evidence set $E$
\State $w \gets \textsc{nfc}(w)$ \Comment{before any offset is computed}
\For{$o \gets 1$ \textbf{to} $2$} \Comment{shortest onset wins}
  \If{$w[0{:}o]$ contains a vowel}
    \State \textbf{continue} \Comment{\textit{um-} prefixes vowel-initial roots}
  \EndIf
  \For{$i \in \{\texttt{um}, \texttt{in}\}$}
    \If{$w[o{:}o{+}2] = i$ \textbf{and} $|w| \geq o+|i|+3$}
      \State $\rho \gets w[0{:}o] \cdot w[o{+}|i|{:}]$
      \If{$\rho \in E$}
        \State \Return $[\,\rho,\ \texttt{<INF:}i\texttt{@}o\texttt{>}\,]$
      \EndIf
    \EndIf
  \EndFor
\EndFor
\State \Return concatenative strip, or $[w]$ if none applies
\end{algorithmic}
\end{algorithm}

The tokenizer has two stages, of which the first is the intervention.

\paragraph{Stage 1: morphological pre-segmentation.} A surface word maps to a
sequence of morphs by Algorithm~\ref{alg:stage1}. For an infixed form the root is
emitted as a \textit{contiguous} unit followed by a positional infix token, so
that \texttt{sumulat} becomes \texttt{["sulat", "<INF:um@1>"]}. The root is now
contiguous in the token stream although discontinuous in the string, which is
precisely what no sequence of BPE merges achieves. Root-then-infix ordering keeps
the word-boundary marker on the root and makes voice a discrete decision the
decoder takes after the root. For Indonesian, a \texttt{NUM CLF N} or
fused \texttt{seCLF N} sequence binds classifier and noun into a single span morph.

\paragraph{Stage 2: constrained BPE over morphs.} Merges are learned by frequency
\textit{within} morphs and can never cross a morph boundary, because the trainer
sees morphs as pre-tokenized units. Granularity therefore remains learned and
corpus-driven; only the positions at which cuts may fall are constrained.

\paragraph{Invertibility.} Exact reassembly is a hard requirement, since decoding
is meaningless without it, and the positional offset in the infix token is what
makes it exact. Section~\ref{sec:defects} reports a defect this requirement
caught.

\subsection{The matched baseline}

The baseline is BPE trained on the \textit{same} corpus lines, with the
\textit{same} word-boundary convention and normalisation and its vocabulary size
chosen to hit the same fertility target, implemented over the same tokenizer back
end
\citep{kudo-richardson-2018-sentencepiece}. It is not an off-the-shelf tokenizer:
whether a merge respects a morpheme boundary is decided by corpus frequency alone,
which is the failure mode under test.

\subsection{The scrambled-morpheme control}
\label{sec:scramble}

\begin{algorithm}[t]
\small
\caption{The scrambled control, for a word the analyzer accepted}
\label{alg:scramble}
\begin{algorithmic}[1]
\Require $w$; true analysis $(i^\star, o^\star)$; per-type cache $\kappa$
\If{$w \in \kappa$}
  \State \Return $\kappa[w]$ \Comment{one assignment per type, corpus-wide}
\EndIf
\State $\mathcal{A} \gets \{(o,k) : o,k \in \{1,2,3\},\
   |w| \geq o+k+3,\ (o,k) \neq (o^\star,|i^\star|)\}$
\If{$\mathcal{A} = \emptyset$}
  \State $\kappa[w] \gets (i^\star, o^\star)$ \Comment{too short to derange}
\Else
  \State $(o,k) \sim \mathrm{Uniform}(\mathcal{A})$
  \State $\kappa[w] \gets (w[o{:}o{+}k],\ o)$
\EndIf
\State \Return $[\,w[0{:}o] \cdot w[o{+}k{:}],\
   \texttt{<INF:}w[o{:}o{+}k]\texttt{@}o\texttt{>}\,]$
\end{algorithmic}
\end{algorithm}

The control fires on \textit{exactly} the word types the real analyzer fires on
and always emits exactly two morphs, but it excises a pseudo-infix at an (offset,
length) pair guaranteed to differ from the true analysis
(Algorithm~\ref{alg:scramble}). The assignment is cached per type, so it is
deterministic given the seed and stable across the corpus. Both arms use the same positional
marker schema, \texttt{<INF:\textit{string}@\textit{offset}>}, so the control
introduces no marker category of its own, although the string a given marker
carries differs between the arms.

\paragraph{Worked example.} For \texttt{sumulat}, $\mathcal{A}$ contains
$(1,1),(1,3),(2,1),(2,2)$ and $(3,1)$, giving the residues \texttt{smulat},
\texttt{slat}, \texttt{suulat}, \texttt{sulat} and \texttt{sumlat}. For
\texttt{sinulat} the same offsets give \texttt{snulat}, \texttt{slat},
\texttt{siulat}, \texttt{silat} and \texttt{sinlat}. The two residue sets meet
only at \texttt{slat}, so of the 25 joint assignments exactly one lets the
paradigm share a unit, and that unit is not the root. For
\texttt{kumain}/\texttt{kinain} the residues are disjoint, so no assignment
collapses the paradigm.

The residue is therefore not a real root, and paradigm members stop collapsing
onto a shared unit. What the control preserves is granularity (morph counts are
identical line by line, verified in Appendix~\ref{sec:appendix-gates}),
invertibility, and per-type consistency of the decomposition. What it destroys is
the intended morphological alignment: the substring excised, its length and
offset, and the residue all change, so the control is not a correctness-only
manipulation.

\subsection{What is tokenized, and where vocabulary parameters enter}
\label{sec:scope}

A single tokenizer is trained over both sides of the bitext and shared by the
encoder and the decoder; the embedding matrix is shared and the output projection
tied to it, so vocabulary parameters enter the count exactly once. Stage-1
segmentation therefore applies to the English source as well as the target, but
the root evidence that licenses an infix analysis is induced from target-side text
alone, so an English word is decomposed only if its residue is an attested Tagalog
type. Vocabulary matching refers to this single joint vocabulary.

\subsection{Fertility matching, and what it costs}
\label{sec:fertility}

A typology-aware tokenizer changes how fine it cuts and where it cuts at the same
time, and only a fertility-matched comparison separates the two. We match on
tokens per character rather than tokens per word, because classifier-span binding
merges two whitespace words into one unit and would drive tokens per word below
1.0 as a pure artifact.

Target levels are derived from the measured intersection of each condition's
achievable envelope rather than fixed a priori, since a target reachable for one
arm can be unreachable for another. The derived levels are 0.319, 0.379 and 0.438
tokens per character, and no cell deviates from its target by more than 0.005, the
tolerance the matching gate asserts.
Appendix~\ref{sec:appendix-fertility} gives the achieved value and vocabulary size
for every cell.

Matching fertility forces different vocabulary sizes. MorphAustro-Tok requires
33.2\%, 19.8\% and 16.3\% more vocabulary entries than BPE to reach the same
granularity at the three levels, since morph-constrained merges buy less
compression per entry. We report this as a result in its own right
(\S\ref{sec:vocab}) and treat the consequent parameter difference as a confound to
quantify rather than a defect to forbid.

\subsection{Two sources of uncertainty, and the correction families}
\label{sec:uncertainty}

\textit{Model-level uncertainty} is variation across seeds, reported as
per-condition means, standard deviations and $t$-intervals. At $n=3$
($t_{\text{crit}} = 4.30$) these intervals are wide, and we treat them as
descriptive.

\textit{Test-example-level uncertainty} is variation attributable to the
evaluation sample with trained models held fixed, captured by a two-sided paired
bootstrap over matched test examples with 10,000 resamples
\citep{koehn-2004-statistical,efron-tibshirani-1993-introduction}. Per-example
scores are averaged across seeds before resampling, so the bootstrap measures
example-level variance and does not re-count seed variance.

Holm correction \citep{holm-1979-simple} is applied within two families fixed in
advance: the 22 contrasts of each non-baseline condition against matched BPE,
pooled across both languages, all tasks and all fertility levels; and, separately,
the nine MorphAustro-Tok versus scrambled contrasts. Pooling the first family
across tasks is conservative, since it corrects the MT contrasts for the NER and
probe comparisons as well.

A bootstrap $p$-value here is a tail proportion over the 10,000 resamples, so the
procedure resolves nothing below $10^{-4}$ before correction. Contrasts whose
uncorrected tail proportion is zero are reported as reaching that floor rather
than with a more precise figure.

Neither source of uncertainty subsumes the other and we never combine them. A
bootstrap interval excluding zero is not evidence that an effect would survive
retraining. We therefore apply a prespecified stability criterion, under which a
claim about mechanism requires both example-level significance and a seed-level
difference exceeding between-seed dispersion. One standard deviation over three
seeds is a conservative heuristic rather than an inferential test.
Section~\ref{sec:mt} reports a case of divergence between the two, and the
divergence is itself the finding.

\section{Implementation}
\label{sec:impl}

\subsection{Corpora and provenance}

Training bitext is OPUS ParaCrawl v9 (\texttt{en-tl}, \texttt{en-id}), fetched
from pinned versioned URLs with SHA-256 recorded over the downloaded bytes; a
failed fetch raises rather than substituting. Filtering (deduplication, length
ratio, language identification) retained 90.0\% of Tagalog and 70.8\% of
Indonesian pairs and is applied identically to every condition. Evaluation uses
FLORES+ (\texttt{fil\_Latn}, \texttt{ind\_Latn}, \texttt{eng\_Latn}) and WikiANN.
Translation quality is measured with \chrf
\citep{popovic-2015-chrf,popovic-2017-chrf}. Appendix~\ref{sec:appendix-config}
gives the training configuration and Appendix~\ref{sec:appendix-data} the corpus
sizes and the sources we rejected.

\paragraph{Contamination.} FLORES+ test text was checked against training bitext
by normalised exact match, character 5-gram Jaccard, and containment. Containment
matters independently: a training sentence that is a \textit{prefix} of a longer
test sentence has low Jaccard while leaking entirely. Measured
contamination was 0.0\% on both languages by all three criteria.

\paragraph{A register gap we could not close.} FLORES+ labels Tagalog as Filipino
(\texttt{fil\_Latn}), the standardised national register, while ParaCrawl is
labelled Tagalog. Infixation is identical across both, so the mechanism is
unaffected, but the train/test gap is large: type-Jaccard is 0.179, with 60.9\% of
test types unseen in training. It applies identically to all conditions and cannot
manufacture a between-condition difference, although it holds absolute scores at
floor level.

\subsection{The mechanism is verifiably active}
\label{sec:activity}

On held-out infixed forms whose roots the corpus attests, segmentation recovers
root and infix in 11 of 11 cases, and on a negative fixture of words that merely
look infixed (\textit{bumbilya} `light bulb', \textit{lumpia} `spring roll')
precision is 1.000, so the analyzer does not fire on everything.
The mechanism fires on 33.75\% of Tagalog lines and 1.63\% of whitespace words,
and MorphAustro-Tok separates from matched BPE by 6.94\% in tokens per character.
Section~\ref{sec:gating} gives the much lower Indonesian figures and explains why
that arm is exploratory.

\subsection{The harness, and the defects it caught}
\label{sec:defects}

The credibility of a control-based null depends on the reader believing that the
null was not produced by a defect. Every validation criterion is runnable code
that halts on failure, and Appendix~\ref{sec:appendix-gates} lists the gates and
the defects they caught. One defect shaped the design. Stage 1 originally computed
the infix offset on the raw surface string while the tokenizer normalised to NFC
afterwards, so where a code point before the offset was decomposed, reassembly
inserted the infix at the wrong position and produced plausible-looking wrong
words rather than errors. The roundtrip gate caught it at a rate of 1 line in
2,000, and a fixture-based check would not have. Normalisation now precedes any
offset computation, which is why it is the first line of
Algorithm~\ref{alg:stage1}.

\subsection{Capability gating, and one arm we treat as exploratory}
\label{sec:gating}

A component whose requirement is unmet is recorded as \texttt{SKIPPED} with a
reason rather than run against a substitute. Malay and Indonesian cannot be
pooled, so Indonesian runs alone
(Appendix~\ref{sec:appendix-indonesian}). Tagalog UD tagging is unusable, since UD\_Tagalog-TRG
contains 55 sentences and Ugnayan 94 \citep{nivre-etal-2020-universal}; WikiANN
\texttt{tl} is substituted, and the substitution is logged rather than silent. The
top rung of the supervision ladder does not exist either, because no maintained
Philippine morphological analyzer is available off the shelf. Rungs 1
(corpus-derived root evidence) and 2 (lexicon) are implemented, and rung 1 alone
attested 9 of 10 held-out fixture roots.

\paragraph{Indonesian is a feasibility arm.} The classifier mechanism reaches
3.95\% of lines and 0.27\% of words, giving a fertility separation of 0.167\%
against 6.94\% for Tagalog, which is too rare to move corpus-level scores at any
seed count this budget affords. We therefore cut Indonesian MT and NER on measured
evidence and kept only the classifier-agreement probe, which is scored on
classifier-bearing sentences and so is unaffected by corpus-level rarity.
Appendix~\ref{sec:appendix-indonesian} gives the detail.

\subsection{A circular probe, measured rather than assumed}
\label{sec:circular}

The Tagalog voice probe is solvable by reading the \texttt{<INF:\dots>} token
alone: a trivial baseline that inspects only which infix token appears scores
1.000 under MorphAustro-Tok and 0.000 under matched BPE. Attempts to break the
circularity all failed for one structural reason, since any corruption of the
infix also changes whether the analyzer fires, which is itself visible in the
token stream. We therefore report the probe as diagnostic rather than evidential;
it carries no evidential weight. The Indonesian probe is not circular in the same
way (Appendix~\ref{sec:appendix-indonesian}).

\section{Evaluation}
\label{sec:eval}

\subsection{Setup}
\label{sec:setup}

\begin{table}[t]
\centering\small
\setlength{\tabcolsep}{3.5pt}
\begin{tabular}{lccc}
\toprule
\textbf{Level} & \textbf{Morph} & \textbf{BPE} & \textbf{$\Delta$ [95\% CI]} \\
\midrule
F0 (0.319) & 11.35 & 10.44 & \textbf{+0.844} [0.682, 1.011] \\
           & \small$\pm$2.18 & \small$\pm$2.27 & \\
F1 (0.379) & 10.48 & \phantom{0}9.29 & \textbf{+1.488} [1.291, 1.690] \\
           & \small$\pm$2.53 & \small$\pm$2.37 & \\
F2 (0.438) & 10.93 & \phantom{0}7.48 & \textbf{+3.396} [3.213, 3.580] \\
           & \small$\pm$1.34 & \small$\pm$1.58 & \\
\bottomrule
\end{tabular}
\caption{\chrf on FLORES+ devtest, mean $\pm$ sd over three seeds, with the
paired-bootstrap difference and its 95\% interval. All three reach the bootstrap
resolution floor of \S\ref{sec:uncertainty}.}
\label{tab:mt}
\end{table}

Three conditions (MorphAustro-Tok, matched BPE, scrambled control) were run at
three fertility levels with three seeds each. The scrambled control is absent at
the finest level, a budget-driven omission. The 33 runs are 24 Tagalog and 9
Indonesian. A Tagalog run trains three
separate models on one tokenizer build and one seed, an encoder--decoder for MT,
an encoder for NER and an encoder for the voice probe, and an Indonesian run
trains the classifier probe alone. Tasks within a run share a configuration
fingerprint, which is what makes the task-transfer correlation well posed.

Model capacity is restricted deliberately rather than by necessity. At between
1.12M and 1.92M inference-path parameters there is little room for a model to
memorise its way around a segmentation that fragments a paradigm, so a difference
between tokenizers surfaces in the score instead of being absorbed by capacity.
We therefore treat the setup as a glass-box diagnostic rather than a competitive
benchmark, and read the absolute scores as a floor.

The in-domain pool is a seed-fixed random sample of the filtered bitext, drawn
before the training split so the two are disjoint. The scored subset is the first
$N$ sentences in corpus order (500 of 1,012 FLORES+ devtest; 300 of 2,000
in-domain), identical under every condition and seed, so the bootstrap pairing is
exact.

NER uses WikiANN \texttt{tl} (10,000 train and 1,000 test sentences, seven IOB2
labels over PER, ORG and LOC), scored with entity-level precision, recall and F1
from \texttt{seqeval}. The tagger is a two-layer Transformer encoder of the same
geometry as the MT encoder, trained for 800 steps with a linear classification
head. A word's representation is the \textit{unweighted mean} of every subword
piece that covers it, obtained by multiplying the encoder states by a
row-stochastic piece-to-word matrix; a word whose pieces are all truncated away
has an all-zero row and is masked out of the loss rather than scored. Labels are
never assigned to the first subtoken, which is not comparable across conditions.

\subsection{Machine translation against matched BPE}
\label{sec:mt}

Table~\ref{tab:mt} reports the comparison against the matched baseline.
MorphAustro-Tok improves \chrf at all three levels, by 0.844, 1.488 and 3.396 mean
sentence-level points, each at the bootstrap resolution floor, and the in-domain
ParaCrawl split reproduces the pattern with larger effects (+1.168, +1.631,
+3.933). Two estimands appear in that table and they do not subtract. The
condition columns are corpus-level scores averaged over seeds, whereas $\Delta$ is
the mean per-sentence difference over the matched evaluation set, since a corpus
aggregate cannot be resampled over examples. The corpus-level gaps are +0.913,
+1.191 and +3.446.

Seed dispersion qualifies two of the three. At F0 and F1 the effect (0.844, 1.488)
lies below the largest between-seed standard deviation in the contrast (2.265,
2.527), and therefore inside retraining noise. Only at F2 does the effect (3.396)
exceed the floor (1.577). By the stability criterion of \S\ref{sec:uncertainty},
F0 and F1 are example-level results that we cannot claim would survive retraining.

\subsection{No detectable contribution from correct boundaries}
\label{sec:scramble-results}

\begin{table*}[t]
\centering\small
\begin{tabular}{llccccc}
\toprule
\textbf{Level} & \textbf{Evaluation set} & \textbf{Morph $-$ BPE} & \textbf{Scrambled $-$ BPE} &
\textbf{Morph $-$ Scrambled} & \textbf{$p_{\text{Holm}}$} & \textbf{Seed sd} \\
\midrule
F1 & FLORES+ devtest  & +1.488 & +1.543 & $-$0.055 & ns & 3.040 \\
F1 & In-domain split  & +1.631 & +1.828 & $-$0.196 & ns & 2.528 \\
F2 & FLORES+ devtest  & +3.396 & +2.735 & +0.661 & at floor & 2.006 \\
F2 & In-domain split  & +3.933 & +3.294 & +0.639 & at floor & 1.566 \\
\bottomrule
\end{tabular}
\caption{Mean per-sentence \chrf differences over the matched evaluation set. The
last column is the pooled between-seed standard deviation for the Morph versus
Scrambled contrast.}
\label{tab:scramble}
\end{table*}

Table~\ref{tab:scramble} reports the study's principal finding. At the central
fertility level the scrambled control matches or exceeds the real analysis on both
evaluation sets, and at the coarsest level the real analysis leads by 0.661 \chrf,
which clears the example-level floor but amounts to one third of between-seed
dispersion. In the table, ``at floor'' marks a bootstrap tail proportion at the
resolution limit of \S\ref{sec:uncertainty}.

Two levels, opposite signs, both far below the retraining floor. By the stability
criterion of \S\ref{sec:uncertainty}, a contribution from linguistic correctness
is not established at either level. That is an absence of detection rather than
equivalence: we specified no margin within which the arms would count as equal,
and a difference below our floor is not the same as no difference.

\subsection{Token count and capacity do not explain the gain}
\label{sec:confounds}

\begin{table}[t]
\centering\small
\setlength{\tabcolsep}{2.5pt}
\begin{tabular}{llrrr}
\toprule
Level & Cond. & Vocab & Params & \chrf \\
\midrule
\multirow{2}{*}{F0} & BPE   & 5{,}842 & 1{,}673{,}984 & 10.44\\
                    & Morph & 7{,}780 & 1{,}922{,}048 & 11.35\\
\midrule
\multirow{3}{*}{F1} & BPE   & 2{,}451 & 1{,}239{,}936 & \phantom{0}9.29\\
                    & Morph & 2{,}936 & 1{,}302{,}016 & 10.48\\
                    & Scr.  & 3{,}567 & 1{,}382{,}784 & 10.62\\
\midrule
\multirow{3}{*}{F2} & BPE   & 1{,}483 & 1{,}116{,}032 & \phantom{0}7.48\\
                    & Morph & 1{,}725 & 1{,}147{,}008 & 10.93\\
                    & Scr.  & 1{,}872 & 1{,}165{,}824 & 10.24\\
\bottomrule
\end{tabular}
\caption{Vocabulary size, inference-path parameters and \chrf for every Tagalog
cell of the sweep.}
\label{tab:capacity}
\end{table}

\paragraph{Not token count alone.} The effect grows as segmentation coarsens,
from +0.84 to +1.49 to +3.40 across F0, F1 and F2. Holding tokens per character
equal does not remove the effect; it increases it. The anti-hypothesis, that any
gain reduces to fewer tokens per character, is not supported.

\paragraph{Not the capacity that fertility matching buys.} Matched fertility
forces MorphAustro-Tok to a larger vocabulary and therefore to more embedding
parameters, so the confound is real and we measured it
(Table~\ref{tab:capacity}). Every count there is $926{,}208 + 128\,|V|$, because
the non-embedding block is identical across conditions and the embedding is
counted once: source and target share one vocabulary and the output projection is
tied to it. The cleanest evidence is a single within-level comparison rather than a
trend across levels. At F2 the scrambled arm carries 18,816 more parameters than
MorphAustro-Tok, exactly the 147 pseudo-infix rows the scramble discovers, and
scores 0.69 \chrf lower, which is the sign a capacity account forbids. Across levels the parameter advantage is
largest where the gain is smallest, which points the same way.

\subsection{A vocabulary cost}
\label{sec:vocab}

MorphAustro-Tok requires 33.2\%, 19.8\% and 16.3\% more vocabulary entries than
matched BPE to reach the same tokens per character, because constrained merges
cannot cross morph boundaries and each entry therefore buys less compression.
Typology-aware segmentation is not free at matched granularity, and the cost is
largest when compression is hardest.

\subsection{Named entity recognition}
\label{sec:ner}

\begin{table}[t]
\centering\small
\setlength{\tabcolsep}{2pt}
\begin{tabular}{llccc}
\toprule
Level & Condition & P & R & F1\\
\midrule
\multirow{2}{*}{F0}
 & BPE       & $.599 \pm .031$ & $.700 \pm .027$ & $.646 \pm .030$\\
 & Morph     & $.601 \pm .023$ & $.703 \pm .009$ & $.648 \pm .017$\\
\midrule
\multirow{3}{*}{F1}
 & BPE       & $.597 \pm .030$ & $.705 \pm .018$ & $.646 \pm .025$\\
 & Morph     & $.618 \pm .018$ & $.720 \pm .008$ & $.665 \pm .013$\\
 & Scrambled & $.587 \pm .008$ & $.694 \pm .010$ & $.636 \pm .009$\\
\midrule
\multirow{3}{*}{F2}
 & BPE       & $.602 \pm .012$ & $.704 \pm .012$ & $.649 \pm .012$\\
 & Morph     & $.623 \pm .020$ & $.719 \pm .010$ & $.667 \pm .015$\\
 & Scrambled & $.616 \pm .011$ & $.715 \pm .009$ & $.662 \pm .004$\\
\bottomrule
\end{tabular}
\caption{WikiANN Tagalog entity-level precision, recall and F1 (\texttt{seqeval}),
mean $\pm$ sd over three seeds.}
\label{tab:ner-raw}
\end{table}

Table~\ref{tab:ner-raw} gives the raw scores. The scrambled arm is the only
condition that falls below the baseline, and it does so on both precision and
recall. The paired-bootstrap differences, which are again sentence-level means and
do not subtract from the corpus-level scores, tell a more specific story. At F1, MorphAustro-Tok does not beat the baseline
(+0.013, $p_{\text{Holm}} = 0.541$) and the scrambled control falls below it
($-0.010$, $p_{\text{Holm}} = 0.919$), while the contrast between the two is
+0.023 at $p_{\text{Holm}} = 0.0045$ against a between-seed pooled sd of 0.011,
the only non-circular result in the study that clears both floors.

The defensible statement is therefore that linguistically wrong morpheme
boundaries damage sequence labelling, not that morphology-aware boundaries improve
it. The effect does not replicate at F2 (+0.003, $p_{\text{Holm}} = 0.850$) and is
absent at F0, so we report it at one fertility level rather than as a general
effect.

\subsection{The Indonesian exploratory arm}

On the classifier-agreement probe, MorphAustro-Tok scores $0.903 \pm 0.003$,
matched BPE $0.900 \pm 0.006$ and the scrambled control $0.896 \pm 0.002$. The
ordering is correct and the differences are negligible; the largest, +0.007
against the scrambled control, is not significant after correction
($p_{\text{Holm}} = 0.168$). We draw no inference from the arm and report it as a
feasibility check only (Appendix~\ref{sec:appendix-indonesian}).

\subsection{No task transfer}

Correlating per-configuration MT and NER gains over the nine shared Tagalog
configurations, joined on run fingerprint so that both tasks provably used the
same tokenizer build and seed, gives Pearson $r = -0.366$ ($p = 0.333$) and
Spearman $\rho = -0.450$ ($p = 0.224$). An MT gain does not predict an NER gain,
which is consistent with reports that tokenizer quality transfers unevenly across
tasks \citep{rust-etal-2021-good}.

\section{Discussion}
\label{sec:discussion}

The natural reading of our results is that a morpheme-boundary constraint on BPE
improves machine translation for Tagalog without requiring the boundaries to be
\textit{correct}.

Both MorphAustro-Tok and its scramble impose the same structural change on the
same set of word types. A word that would otherwise be segmented by whatever the
merge table favours is instead always decomposed the same way, everywhere it
occurs. Frequency-driven BPE has no such guarantee, since a rare inflected form
may be cut differently from a frequent one. Both conditions remove that
inconsistency.

Consistency is the property the two arms share that the baseline lacks, and it
remains the leading candidate, but the design does not isolate it from what the
arms also share: the same triggering types, morph count and root-plus-token
shape.

If the reading is right, it has a practical implication that runs against the
motivation for this class of intervention: the expensive part may not be the part
that pays. Morphological analyzers are costly and scarce for exactly the languages
that most need them, and a linguistically arbitrary decomposition captured the
same MT gain here.

\paragraph{Threats to validity.} The scrambled control is not linguistically
random in every respect: it fires on the same word types the real analyzer
selects, so it inherits the decision about \textit{which} words to decompose while
randomising \textit{how}. The selection of infixed types is itself linguistic
information and may carry the effect, in which case the conclusion narrows further
to ``correct placement does not matter given correct selection''. That is the most
important open question our design leaves. Scale is a second threat: an effect
visible only once a model can exploit paradigm structure would be invisible at the
size we trained.

\paragraph{The sequence-labelling asymmetry.} The NER result sits awkwardly with
the MT result and we do not have a confident account of it. Sequence labelling
appears sensitive to wrong boundaries in a way that generation is not: on NER,
scrambling costs 0.023 F1 relative to real morphology, while on MT it costs
nothing we can detect. One possibility is that word-level label pooling makes
token-internal structure matter more than it does for generation, whose decoder
can compensate. The result replicates at only one of two levels,
and we flag it as a lead rather than a finding.

\section{Conclusion}

Across 33 runs, a morpheme-boundary constraint on BPE improved Tagalog machine
translation over a matched baseline at all three fertility levels, by 0.84, 1.49
and 3.40 mean sentence-level \chrf points. Token count does not explain the gain,
since it grows as segmentation coarsens, and neither does the capacity that
fertility matching buys: at the coarsest level the arm with more parameters scores
lower.

A scrambled control preserving granularity, invertibility and per-type consistency
but not morphological validity matched or exceeded the real analysis at the
central fertility level and trailed it by one third of a standard deviation at the
coarsest. Sequence labelling differs: scrambling costs 0.023 F1 at one level,
although MorphAustro-Tok still does not beat the baseline.

In conclusion, the claim we defend is narrow: for this language, at this scale, we
detect no contribution from segmenting correctly, and the result is consistent
with a role for segmenting consistently that a further control would have to
confirm. Three follow-ups are implied: a
stronger scramble that randomises type selection, a consistency-only baseline,
and more seeds.

\section*{Limitations}

First, scale. Small models are a design choice rather than a constraint we worked
around (\S\ref{sec:setup}), but the choice has a cost. The fertility-matched models
carry between 1.12M and 1.92M inference-path parameters and are trained from
scratch for 1,500 steps, roughly 1.6 epochs, so every model is under-trained. The constraint applies
identically across conditions and the contrasts therefore hold, but absolute
scores are at floor level (\chrf 6--11) and we cannot claim that these effects
persist at contemporary scale or under transfer from a pretrained checkpoint.

Second, three seeds. With $t_{\text{crit}} = 4.30$, seed-level intervals are wide
and strictly descriptive. Any effect smaller than between-seed dispersion is
undetectable here, and two of our three MT effects fall in that range.

Third, there is no scrambled control at the finest fertility level. F0 was run
with two conditions only, for budget reasons, so the F0 MT effect is uncontrolled
and we make no attribution for it.

Fourth, the scrambled control inherits type selection. It randomises how a word is
decomposed, not whether it is decomposed at all, so the design narrows the
explanation without isolating consistency.

Fifth, the Indonesian arm is underpowered by construction rather than by accident
of sample size, since the mechanism reached 0.27\% of words. It is reported as an
exploratory feasibility check and supports no inference either way.

Sixth, the voice probe is circular and is reported as diagnostic only
(\S\ref{sec:circular}). The evidential weight rests on MT and NER.

Seventh, the register gap. ParaCrawl Tagalog against FLORES+ Filipino leaves
60.9\% of test types unseen in training. The gap is symmetric across conditions,
but it holds absolute quality at floor level and may interact with the effects we
measure in ways we cannot separate.

Eighth, decoding is greedy and evaluation is capped at 500 FLORES+ and 300
in-domain sentences, both for budget. These settings are applied identically to
every condition; the bootstrap runs on a smaller matched set and its intervals are
correspondingly wider.

Ninth, language identification in bitext filtering is weak. It is a function-word
heuristic rather than a trained classifier, and it is unreliable on short
sentences.

Tenth, one line in 2,000 failed exact roundtrip under the morphological conditions
after the normalisation fix. The line is a training sentence rather than an
evaluation example, so no paired comparison is affected, and the exclusion is
applied to the union over conditions, so no condition is trained on material the
others lack.

Eleventh, the study covers two languages from one family and one tokenizer
comparison. A unigram-LM tokenizer \citep{kudo-2018-subword} at matched vocabulary
is the obvious missing baseline.

\section*{Ethical Considerations}

This work uses publicly released corpora under their stated licences. FLORES+ is
distributed under CC-BY-SA-4.0 and is gated, so access requires accepting its
conditions; ParaCrawl is released under CC0; and WikiANN is released under
ODC-BY-1.0. We ask that work building on this paper cite the originating datasets
and their contributors.

ParaCrawl is web-mined and is known to contain noise and material its authors did
not intend for redistribution, and our filtering removes neither personal
information nor offensive content in any targeted way. We did not audit the
Tagalog or Indonesian portions for either, and we release no model weights trained
on them.

Two of the languages studied here are under-served by language technology, and
work of this kind carries a risk of overclaiming on their behalf. Our central
result is a negative one, and we have tried to report it in terms that do not
invite the opposite reading: a scrambled control performing as well as a real
morphological analysis is evidence about our experimental scale, not evidence that
Tagalog morphology is unimportant or that linguistic expertise is dispensable in
building resources for Philippine languages.

The compute cost of the reported experiments is 0.76 GPU-hours on a single T4. A
further 6.3 GPU-hours were spent on runs discarded after a throughput rescale and
on tokenizer construction. We report the two figures separately rather than as an
aggregate, since folding discarded compute into the reported cost misrepresents
what reproducing this study requires.

\bibliography{custom}

\appendix

\section{Training configuration}
\label{sec:appendix-config}

Table~\ref{tab:config} lists every setting. All of them are held fixed across the
three tokenizer conditions, and only the vocabulary size implied by fertility
matching differs. Parameter parity is enforced at matched vocabulary, where all
three conditions carry 1,950,208 inference-path parameters; under fertility
matching the counts range from 1.12M to 1.92M and are given in
Table~\ref{tab:capacity}.

\begin{table}[h]
\centering\small
\setlength{\tabcolsep}{3pt}
\begin{tabular}{ll}
\toprule
\multicolumn{2}{l}{\textit{Architecture (identical in every condition)}}\\
\midrule
Encoder / decoder layers & 2 / 2\\
$d_{\text{model}}$ & 128\\
Attention heads & 4\\
FFN dimension & 512\\
Dropout & 0.1\\
Normalisation & pre-LN, GELU\\
Positional encoding & sinusoidal, non-learned\\
Source/target embedding & single shared matrix\\
Output projection & tied to input embedding\\
Initialisation & PyTorch defaults\\
\midrule
\multicolumn{2}{l}{\textit{Optimisation (identical in every condition)}}\\
\midrule
Optimiser & AdamW, $\beta=(0.9,0.98)$,\\
 & $\epsilon=10^{-9}$\\
Peak learning rate & $3\times10^{-4}$\\
Schedule & inverse-sqrt, 200 warmup steps\\
Weight decay & 0.01, matrices only\\
Gradient clipping & 1.0 (global norm)\\
Batch size & 32 sentences\\
Gradient accumulation & none\\
Label smoothing & 0.1\\
Mixed precision & AMP fp16 on T4\\
Max source/target length & 64 (MT), 96 (tagging)\\
\midrule
\multicolumn{2}{l}{\textit{Schedule and selection}}\\
\midrule
MT steps & 1{,}500\\
NER steps & 800\\
Probe steps & 400\\
Evaluation cadence & once, at the final step\\
Checkpoint selection & final step, no early stopping\\
Decoding & greedy, batch 32\\
Seeds & 13, 17, 23\\
\bottomrule
\end{tabular}
\caption{Training configuration.}
\label{tab:config}
\end{table}

\section{Data sizes and rejected sources}
\label{sec:appendix-data}

\begin{table}[h]
\centering\small
\setlength{\tabcolsep}{3pt}
\begin{tabular}{lrr}
\toprule
 & Tagalog & Indonesian\\
\midrule
ParaCrawl pairs parsed & 248{,}689 & 400{,}000\\
\quad retained after filtering & 223{,}832 & 283{,}018\\
\quad retention rate & 90.0\% & 70.8\%\\
\midrule
MT train & 30{,}000 & 30{,}000\\
In-domain held-out pool & 2{,}000 & 2{,}000\\
\quad scored & 300 & n/a\\
FLORES+ dev & 997 & 997\\
FLORES+ devtest & 1{,}012 & 1{,}012\\
\quad scored & 500 & n/a\\
\midrule
WikiANN train / dev / test & \multicolumn{2}{r}{10{,}000 / 1{,}000 / 1{,}000}\\
Tokenizer corpus (lines) & 70{,}000 & 72{,}000\\
Root-evidence corpus (lines) & 231{,}832 & 293{,}018\\
Probe examples (80/20 split) & 6{,}000 & 6{,}000\\
\bottomrule
\end{tabular}
\caption{Corpus sizes. WikiANN rows are the Tagalog splits.}
\label{tab:data}
\end{table}

The Indonesian raw count in Table~\ref{tab:data} is a parse cap rather than the
size of ParaCrawl \texttt{en-id}. The tokenizer corpus is both sides of the MT
training bitext plus the NER training sentences. Root evidence is target-side text
only: the retained bitext plus the 10,000 NER training sentences, less the
in-domain held-out pool when part of that pool is scored, which is why the Tagalog
count is 2,000 lower than the Indonesian construction would suggest.

The Indonesian raw count in Table~\ref{tab:data} is a parse cap rather than the
size of ParaCrawl \texttt{en-id}, and the Indonesian WikiANN splits (12,000 /
10,000 / 10,000) were never used, since Indonesian NER was cut. The tokenizer
corpus is both sides of the MT training bitext plus the NER training sentences.
Root evidence is target-side text only: the retained bitext plus the 10,000 NER
training sentences, less the in-domain held-out pool when part of that pool is
scored, which is why the Tagalog count is 2,000 lower than the Indonesian
construction would suggest.

Candidate sources were rejected on recorded grounds.
\texttt{facebook/flores} and \texttt{allenai/nllb} are loading-script datasets and
are unloadable under current tooling. OPUS-100 contains no Tagalog configuration.
The wikimedia, WikiMatrix and translatewiki corpora are Wikipedia-derived and
would contaminate a Wikipedia-sourced test set by construction. XLEnt supplies
entity alignments rather than sentence bitext, and OpenSubtitles has unclear
redistribution terms.

\section{The fertility sweep}
\label{sec:appendix-fertility}

Table~\ref{tab:fertility} gives the achieved value for every cell. The largest
deviation from target is 0.0048, within the 0.005 tolerance the matching gate
asserts. Vocabulary is the size returned by bisection; the scrambled arm
additionally registers 147 discovered pseudo-infix types after training, which is
why its counts in Table~\ref{tab:capacity} are higher by that amount.

\begin{table}[h]
\centering\small
\setlength{\tabcolsep}{5pt}
\begin{tabular}{llrr}
\toprule
Level & Condition & Vocab & Achieved\\
\midrule
\multirow{2}{*}{F0 (target 0.31867)}
 & BPE       & 5{,}842 & 0.31579\\
 & Morph     & 7{,}780 & 0.31732\\
\midrule
\multirow{3}{*}{F1 (target 0.37858)}
 & BPE       & 2{,}451 & 0.38177\\
 & Morph     & 2{,}936 & 0.38335\\
 & Scrambled & 3{,}420 & 0.37490\\
\midrule
\multirow{3}{*}{F2 (target 0.43849)}
 & BPE       & 1{,}483 & 0.43582\\
 & Morph     & 1{,}725 & 0.43537\\
 & Scrambled & 1{,}725 & 0.43803\\
\midrule
\multirow{3}{*}{ind F1 (target 0.37849)}
 & BPE       & 2{,}451 & 0.37395\\
 & Morph     & 2{,}451 & 0.37424\\
 & Scrambled & 2{,}451 & 0.37440\\
\bottomrule
\end{tabular}
\caption{Achieved tokens per character and the vocabulary size each condition
required to reach it.}
\label{tab:fertility}
\end{table}

Across the three fertility levels the parameter advantage of MorphAustro-Tok over
matched BPE and its \chrf gain are ordered oppositely (Pearson $r = -0.810$). With
three points this is descriptive only, and the argument in \S\ref{sec:confounds}
rests on the within-level comparison at F2 instead.

\begin{figure}[h]
\centering
\includegraphics[width=\columnwidth]{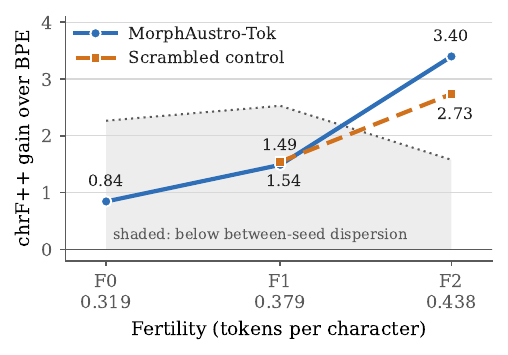}
\caption{Gain over matched BPE against fertility, with the shaded region marking
differences smaller than the between-seed standard deviation.}
\label{fig:effect}
\end{figure}

The Tagalog condition matrix is three seeds for each of MorphAustro-Tok and
matched BPE at all three levels, plus three seeds for the scrambled control at F1
and F2, giving 24 runs. The Indonesian arm adds three conditions at one level,
giving nine. Figure~\ref{fig:effect} plots the two MT contrasts across the sweep.

\section{Gates and further defects}
\label{sec:appendix-gates}

Beyond the normalisation defect described in \S\ref{sec:defects}, the gates cover
granularity parity between the real and scrambled segmenters (identical morph
counts on 2,000 of 2,000 lines), derangement rate of the scramble (1.000 over 84
infixed types), marker-codepoint disjointness from corpus text, achieved-fertility
tolerance, vocabulary-count parity, decoder causality (maximum logit difference
0.00 at positions preceding the cut, asserted on every run), exact roundtrip, and
a pooling audit asserting that each row covers exactly its word's pieces.

Three further defects are worth recording. The BPE trainer was seeing the literal
characters of tokens already registered as atomic and learning merges from inside
them, so that fragments occupied real vocabulary slots and were emittable during
decoding. The Indonesian classifier binder handled \texttt{NUM CLF N} but not the
fused \texttt{seCLF N} form that carries most Indonesian classifier constructions,
which left activity at 0.02\% of words and a fertility separation of 0.07\%,
amounting to three labels on one tokenizer. Finally, a single wall-clock ceiling
per run allowed machine translation to consume the entire allowance, leaving NER
and the probe at step 1 of their schedules, and the first live run took 182
minutes against a 12.6-minute estimate, of which roughly 160 minutes was
unbudgeted batch-of-one beam search. Ceilings are now per task, decoding is
batched and greedy, and the affected runs were excluded after the matrix was
rescaled.

\section{The Indonesian feasibility arm}
\label{sec:appendix-indonesian}

Malay and Indonesian cannot be pooled, which FLORES+ makes directly measurable
rather than assumed: over the 997 sentence pairs present in both
\texttt{zsm\_Latn} and \texttt{ind\_Latn}, 0.0\% are identical, mean normalised
edit distance is 0.425 and type-Jaccard is 0.391. Indonesian therefore runs alone.

The classifier binder originally handled \texttt{NUM CLF N} but not the fused
\texttt{seCLF N} form that carries most Indonesian classifier constructions, which
left activity at 0.02\% of words and a fertility separation of 0.07\%, amounting
to three labels on one tokenizer. After the fix the mechanism reaches 3.95\% of
lines and 0.274\% of words, with a fertility separation of 0.167\%. That is two
orders of magnitude below the Tagalog separation of 6.94\%, so Indonesian MT and
NER were cut on measured evidence rather than attempted and reported as null.

What remains is the classifier-agreement probe, a corruption-judgment task whose
label is whether a classifier and its noun were swapped. No tokenizer exposes that
label, so the probe is not circular in the way the Tagalog voice probe is, and the
trivial token-inspection baseline scores 0.000 on it. Against matched BPE
MorphAustro-Tok gains +0.003 ($p_{\text{Holm}} = 0.919$), and against the
scrambled control +0.007 ($p_{\text{Holm}} = 0.168$). We read the arm as evidence
that the mechanism can be built and exercised in a second Austronesian language,
not as evidence about classifier-aware segmentation.

\section{Reproducibility}
\label{sec:appendix-repro}

Corpora are fetched at run time from pinned versioned URLs and hub identifiers,
with SHA-256 recorded over materialised content, and with licences and citations
recorded in the run manifest; a failed fetch raises rather than substituting. We
release at \url{\coderepo} the notebook that implements the pipeline end to end,
per-run JSON records with metrics and per-example scores, the full gate log, the
dispersion and comparison tables in CSV, and the limitations log.

Seeds are fixed and recorded, and string hashing uses SHA-256 rather than Python's
salted hash. Runs are keyed by architecture and configuration fingerprints, so a
stale artefact is invalidated rather than silently reused. Dependency constraints
are satisfiable version floors verified after import, with resolved versions
recorded; exact pins were unsatisfiable on the runtime available and would have
been recorded as applied while silently failing.

\paragraph{Determinism.} Deterministic algorithms are enabled in PyTorch in
warn-only mode. On the available hardware, fp16 reductions under automatic mixed
precision have no deterministic implementation. We therefore claim seed-controlled
reproducibility of the experimental protocol, not bitwise reproducibility of a
forward pass.

\end{document}